\documentclass[runningheads]{llncs}
\usepackage[T1]{fontenc}
\usepackage{graphicx}
\usepackage{hyperref} 
\usepackage{times}
\usepackage{latexsym}
\usepackage{caption}
\usepackage{microtype}
\usepackage{tabularx}
\usepackage{amsmath, amssymb, mathtools}
\usepackage[capitalize,noabbrev]{cleveref}

\usepackage{color}

\newcommand\bnorm[1]{\bigg\lVert#1\bigg\rVert}

\newcommand\set[1]{\left\{ #1 \right\}}

\begin{document}
\title{Generalized knowledge-enhanced framework for biomedical entity and relation extraction}
%
%
\author{Minh Nguyen \inst{1} \and Phuong Le \inst{2}}
%
\authorrunning{Nguyen, Le}
%
%
\institute{University of Texas at Austin \and University of Illinois Urbana-Champaign}

\maketitle
\begin{abstract}
In recent years, there has been an increasing number of frameworks developed for biomedical entity and relation extraction. This research effort aims to address the accelerating growth in biomedical publications and the intricate nature of biomedical texts, which are written for mainly domain experts. To handle these challenges, we develop a novel framework that utilizes external knowledge to construct a task-independent and reusable background knowledge graph for biomedical entity and relation extraction. The design of our model is inspired by how humans learn domain-specific topics. In particular, humans often first acquire the most basic and common knowledge regarding a field to build the foundational knowledge and then use that as a basis for extending to various specialized topics. Our framework employs such common-knowledge-sharing mechanism to build a general neural-network knowledge graph that is learning transferable to different domain-specific biomedical texts effectively. Experimental evaluations demonstrate that our model, equipped with this generalized and cross-transferable knowledge base, achieves competitive performance benchmarks, including BioRelEx for binding interaction detection and ADE for Adverse Drug Effect identification.

\end{abstract}
\section{Introduction}\label{intro}
A tremendous increase in the number of biomedical publications in recent years \cite{rapid_bio} makes it difficult for biomedical researchers to keep up with latest articles. Consequently, numerous studies such as \cite{extract1,extract2,extract3,extract5,extract6,extract7} in applied deep learning and natural language processing are dedicated towards automatic extraction of biomedical entities and their relations. To validate and enhance these applied machine learning (ML) efforts, several tasks and datasets regarding this topics have also been developed. These include tasks such as: binding interaction detection BioRelEx \cite{Khachatrian2019-ms}, adverse drug effect ADE \cite{Gurulingappa2012-kw}, drug-drug interaction DDI \cite{Herrero-Zazo2013-yf}, and bacteria biotope task BB-rel \cite{Bossy2019-kj}, each targeting specific aspects of biomedical research.\\

\noindent Compared to understanding general text, information extraction of biomedical text documents requires much broader domain knowledge since these documents contain many technical terms usually intended only for domain experts \cite{extract4}. To successfully perform the joint tasks of entity and relation extraction, the resulting models need to understand these terms thoroughly, from their technical definitions, semantic types to how they are connected with others. For example, assume we are given the biomedical input text: {\color{red} The effect was specific to \textbf{rapamycin}, as \textbf{FK506}, an immunosuppressant that also binds \textbf{FKBP12} but does not target \textbf{mTOR}, had no effect on the interaction.} Then the final model should be able to extract the entity: {\color{red}{rapamycin, FK506, FKBP12, mTOR}} and label them as {\color{blue}{Drug, Chemical, Protein, Protein}} accordingly. Here Drug, Chemical, and Protein are entity types. Moreover, the model should extract 3 relations between 3 pairs {\color{red}(rapamycin, FKBP12), (rapamycin, mTOR), (FK506, FKBP12)}, and label all of these 3 relations as Binds, where Binds is the relation type.

\subsection{Related-works}
For domain-specific document understanding, an advanced strategy is to learn not only from the given input texts but also from the external domain knowledge that supports the comprehension of complex terms. The external knowledge is obtained via a secondary source of texts different from the original inputs. Many machine learning models labeled as knowledge-enhanced models have been developed based on this approach. For instance, Peters et al. \cite{Peters2019-ar} develop the KnowBertAttention, a state-of-the-art knowledge-enhanced language model. KnowBertAttention makes use of SciBERT \cite{Beltagy2019-bq} for token-level representations and employs the KAR mechanism to introduce external knowledge from UMLS. Lai et al. \cite{Lai2021-op} proposes a knowledge-enhanced model with collective inference called KECI. Instead of only extracting features as in KnowBertAttention, KECI injects knowledge from UMLS, makes use of multi-relational graph structure of candidate entities, and integrates more global information to the representations. \\

Below, we provide more details on KECI model, outlining the framework's individual steps which typically align with those taken by knowledge-enhanced models:
\begin{enumerate}
    \item The KECI model initially processes task-specific input documents using text embeddings from \cite{Beltagy2019-bq}. It then constructs a knowledge graph using bidirectional Graph Convolutional Networks (GCN). The outputs of this step comprises feature vectors for all relevant biomedical entities. 
    \item Next, the model utilizes external text sources such as UMLS \cite{Bodenreider2004-ip} to establish a background knowledge graph (KG), following a procedure similar to the first step. Here, the outputs also consist of feature vectors for biomedical entities; however, these entities are sourced from UMLS instead of the original input data.
    \item Finally, KECI integrates feature vectors from the task dataset with feature vectors of relevant entities in the external knowledge dataset to make the final predictions.
\end{enumerate}

\subsection{Motivations and contributions}
One potential problem with these approaches is the waste of external knowledge and resources. For instance, in KECI \cite{Lai2021-op}, the training of the background knowledge graph (KG) is based on a loss function that is task-dependent. Thus, this approach requires rebuilding the KG entirely for each new task, despite the knowledge being commonly applicable and shareable across different tasks. Moreover, KECI uses MetaMap \cite{Aronson2010-uy} to extract relevant biomedical entities for a specific task dataset. This means that, for the model to work with various tasks, we have to extract entities across many task datasets. Consequently, the resulting KG can include a significant number of nodes, and not every pair of nodes has a strong associated edge. \\

To address these shortcomings, we introduce a knowledge-enhanced model designed to leverage external general knowledge from various sources more efficiently. Our approach focuses on building a general knowledge graph (KG) that can truly serve as a public knowledge source and can be readily shared with any new domain-specific tasks. In particular, feature vectors associated with terms and relations already existing in this generalized knowledge graph (KG) can be directly applied to specific biomedical domains tasks, eliminating the need for extensive retraining.\\

Similar to KECI and other knowledge-enhanced models, our framework initially processes the input text to construct an initial graph, where nodes represent entity mentions and edges denote their relation types. However, our approach proposes a novel strategy in the subsequent step where we establish a generalized knowledge graph with two distinct components: \textbf{General-Knowledge} (GK) and \textbf{Specific-Knowledge} (SK). Here, the General-Knowledge (GK) component encapsulates reusable domain knowledge that is independent of specific tasks. This component serves as a foundational repository of broad biomedical domain insights, readily applicable across various tasks without the need for task-specific adaptation. In contrast, the Specific-Knowledge (SK) component is task-dependent and is tailored to individual tasks. This component enhances the adaptability and precision of our model to address the specific requirements for different tasks across the biomedical domain. Our specific contributions are as follows:
\begin{enumerate}
    \item \textbf{Task-independence and reusability}: We develop a novel technique to build the General-Knowledge (GK) component with a `graph-like' structure that contains readily-used feature vectors. The GK component serves as a versatile knowledge base that can be shared and reused across different biomedical tasks, promoting efficiency and consistency in model performance.
    \item \textbf{Efficient knowledge utilization}: By separating GK and SK components, we optimize the utilization of general biomedical domain knowledge, minimizing redundancy and enhancing scalability across diverse tasks. 
    \item \textbf{Focused and adaptive modeling}: We introduce a process to fuse GK with SK components to allow effective learning sharing across different biomedical domains. The SK component ensures that our model can adapt dynamically to the unique characteristics and complexities of specific biomedical tasks, improving accuracy and relevance in predictions. 
    \item Finally, we provide competitive benchmarking experiments against several state-of-the-art text-understanding models on specific biomedical domains including \textbf{binding interaction detection (BioRelEx)} and \textbf{adverse drug effects (ADE)}. 
\end{enumerate}

The rest of the paper is organized as follows: the second section gives details about our knowledge-enhanced framework. The third section shows experiments and benchmarking results for applying our framework to biomedical entity and relation extraction tasks. In the fourth section, we have a comprehensive discussion on the effectiveness of our model in terms of relational extraction and the re-usability of our general knowledge component across multiple specific biomedical tasks. We end our paper with a brief conclusion and future works.


\section{Methods}
\subsection{Overview}
In this section, we provide details of our generalized background knowledge graph framework that can then be utilized efficiently across entity and relation extraction tasks.\\

\noindent\textbf{Generalized Background Knowledge.} Our model aims to build a generalized background knowledge graph that allows common knowledge to be shared across tasks. Our graph include two components: \textbf{general-knowledge (GK)} component and \textbf{specific-knowledge (SK)} component. Entities and relations in our general knowledge (GK) component remain unchanged regardless of specific tasks. For GK component, we elevate biomedical language models to build a ‘graph-like’ structure independent of the input data. For the second (SK) component, we process the given input documents for a specific tasks to build relevant neural-network graphs of entities, and then joins this SK with the first GK component.

\subsection{General-Knowledge (GK) Component}\label{part1}
To construct the General-Knowledge (GK) Component, we need to obtain general entities representations encoding both label and relational information without having to build and train GCN. \\

\noindent\textbf{Extracting relational data.} We first start by utilizing BioBERT to extract relational information from entities within the knowledge source data.\\ 

\noindent Suppose we're given a set of relations $\set{r_k}_{k=1}^{R}$, where $R$ is the number of possible relations, and the triplets consisting of subject, relation and object $\set{(s_i,r_k,o_j)}$. We apply BioBert embedding to get a set of associated weights that indicate how likely each relation $r_k$ will match the subject-object pair ($s_i$, $o_j$), where $i$ and $j$ run over all possible indices of the subjects and objects sets.\\

\noindent For this weight extraction, we start with masking techniques to form four possible hypothetical sentences:
\begin{itemize}
    \item Sentence with relation masked: A = $s_i$ \text{[MASK]} $o_j$
    \item Sentence with object masked: B = $s_i$ $r_k$ \text{[MASK]}
    \item Sentence with subject masked: C = \text{[MASK]} $r_k$ $o_j$
    \item Sentence with no mask: D = $s_i$ $r_k$ $o_j$
\end{itemize}

\noindent Such maskings support the inference process of the remaining entity from two known other entities, and hence implicitly extract the relational information. For instance, to fill in the mask in the first sentence (sentence A), the model needs to approximate a function $\hat{r} = g(s, o)$ so that $\hat{r_k} = g(s_i, o_j)$ approximates $r_k$. The function $g$ provides insights into how a relation can fit in a given pair of (subject, object). \\

\noindent Now from such for (masked) sentences $A, B, C, D$, we use a geometric argument to calculate the expected weight indicating how much a relation $r_k$ will match with the pair of (subject, object). First of all, we convert each sentence into an embedding so that we can get corresponding vectors to perform mathematical operations. For this step, we apply BioBert model to obtain the corresponding embeddings $v_A, v_B, v_C, v_D$.\\

\noindent Next, for each sentence above, we apply BioBERT to obtain its vector representation $v_A, v_B, v_C$ and $v_D$ respectively. Mathematically, $v_A$ can be represented as the ordered sum $v(s_i) + v(\hat{r_k}) + v(o_j)$, where $\hat{r_k}$ is the predicted relation to \textit{fill in the blank} of the sentence with subject $s_i$ and $o_j$. Similarly, $v_B = v(s_i) + v(r_k) + v(\hat{o_j})$, and $v_C = v(\hat{s_i}) + v(r_k) + v(o_j)$, where $\hat{s_i}$ and $\hat{o_j}$ are the predicted object and subject given that two other entities are known.\\ 

\begin{figure}[h!]
\centering
\includegraphics[scale=0.5]{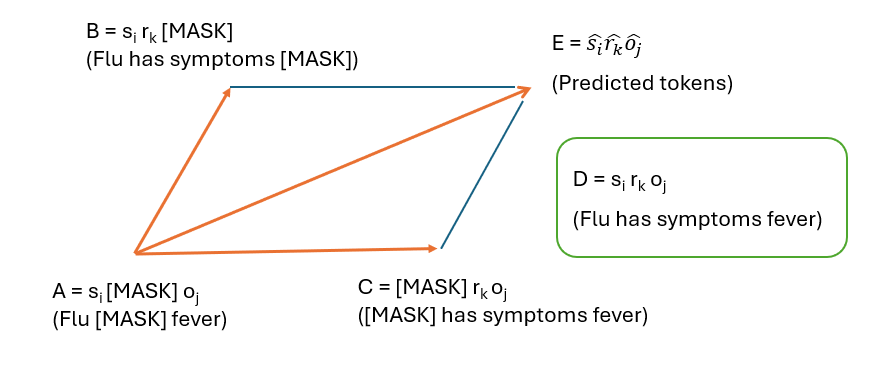}
\caption{Extracting relation data process using BioBERT: (subject, relation, object) = (flu, has symptoms,  fever).}
\label{relation}
\end{figure}

\noindent We then perform the geometric vector operation $v_E = v_B+v_C-v_A$. Here $E$ can be considered as the remaining vertex of the parallelogram with existing vertices $A, B, C$. This geometric operations will yield an approximate embedding of the predicted sentence $E = \hat{s_i}\hat{r_k}\hat{o_j}$ with both predicted subject $\tilde{s_j}$, predicted object $\tilde{o_j}$, and predicted relation $\tilde{r_k}$. The approximation is due to the equation:
\begin{equation}
v_B + v_C - v_A = v(\hat{s_i}) + v(\hat{r_k}) + v(\hat{o_j}) + 2(v(r_k) - v(\hat{r_k}))
\end{equation}

\noindent Now to predict how likely the relation $r_k$ fit in, we calculate the difference between the original sentence $D = s_i r_k o_j$ and its predicted counterpart $E = $ by simply taking the cosine similarity between their embedding: $\hat{w}(s_i, r_k, o_j) = \cos(v_D, v_E)$. See \cref{relation} for an illustration of the above construction.\\

\noindent Here we regard $E$ as BioBERT's `ground truth' on the supposed subject, object, and associated relation (or at least related entities). Thus, when $D$ is compared with $E$, the cosine similarity reflects BioBERT's estimate on how close the hypothesis sentence $D$ is to its prediction. The higher the cosine similarity, the more likely that the relation $r_k$ is correct for pair $(s_i,r_k,o_j)$. Our construction ensures that all 3 components subject, object, and relation are taken into account equally.\\

\noindent\textbf{Entities representation training.} Based on the weights $\hat{w}(s_i, r_k, o_j)$ obtained from previous procedure, we train a feed-forward neural network (FFNN) to encode this relational weights into the entities representations. In particular, we start by taking union over all entities to get a set of distinct entities, each of which is fed into BioBERT model to get its associated initial embedding. Next, we train the FFNN $f$ with the following loss function:
\begin{equation}
L(\theta) = \sum \limits_{s_i} \sum \limits_{k=1}^{R} \bnorm{f(s_i) - \sum \limits_{o_j}\frac{\hat{w}(s_i, r_k, o_j)}{\sum_j \hat{w}(s_i, r_k, o_j)}f(o_j)}_2^2
\end{equation}
where $f(s_i)$ and $f(o_j)$ is the output vector representations of subject $s_i$ and object $o_j$ that take relation weights into account. The final embedding for each entity is a concatenation of the initial embedding from BioBERT and the (relational) embedding $f(.)$ obtained from this training.

\subsection{Specific-knowledge (SK) component and fusion with GK component}\label{part2}
\begin{figure}[h!]
    \centering
    \includegraphics[scale=0.46]{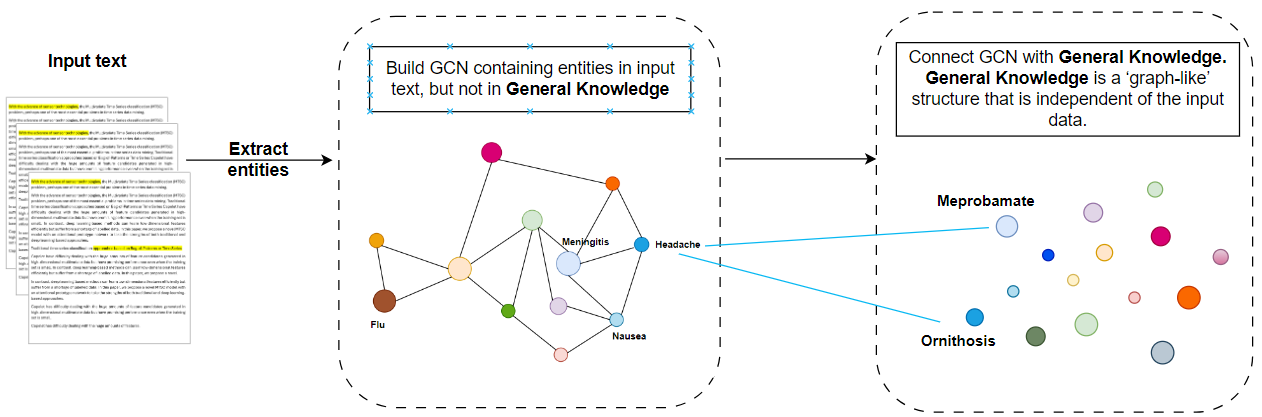}
    \caption{Illustration of building specific task’s graph and connecting with GK.}
  \label{connectGK}
\end{figure}

For the SK component, we follow the standard procedure to obtain a graph neural network between entities from the training input documents for a specific task. More specifically, we obtain all possible entities from given input documents. We build a graph convolutional network (GCN) \cite{gcn} on the entities excluding those in GK component (from \cref{part1}). At this point, we have a GCN containing information specific to a task, together with the reusable GK component containing "general" entity encodings that already carry their relation information. To perform the fusion, we train an additional GCN that connects the nodes in the specific task's graph to those in the graph from the GK component. No edge in the GK component need to be included in this additional fusion GCN. The relations of entities in the GK component are, in fact, independent of the downstream tasks. \cref{connectGK} above presents a simplified pipeline for this process.

\section{Experiments}
\subsection{Data sources and tasks}
\noindent\textbf{Sources data.} We use two data sources for the first GK component in our framework. The first data source is UMLS, which  includes Metathesaurus and Semantic Network. Metathesaurus provides information about millions of biomedical concepts and relations between them. By using MetaMap, the entity mapping tool for UMLS, we can extract relevant UMLS biomedical entities from any input document. Semantic Network, together with Metathesaurus, will then provide relevant relations between those biomedical entities. We note that the UMLS used in \cref{part1} is a simplified and processed version from \cite{Sung2021-cd}, where the author investigates whether LMs can be used as biomedical KB. We group the relations by similarity and further reduce to a total of 5 relation properties including: drug used for treatment, physiologic effect, has symptoms, clinically associated, and drug agent. The second data source we look at is Wikidata \footnote{https://www.wikidata.org}, a public knowledge base with items across domain. The version of Wikidata used, which contains only biomedical entities and relations, is also from \cite{Sung2021-cd}.\\

\noindent\textbf{Tasks.} We evaluate our framework using two biomedical datasets: BioRelEx and ADE. The ADE data contains 4272 sentences (from medical reports) that describe drug-related adverse effects. ADE has 2 entity types (Adverse-Effect and Drug) and a single relation type (Adverse-Effect).  The BioRelEx is a collection of biomedical literature that capture binding interactions between proteins and/or biomolecules. It has 2010 sentences, 33 entity types, 3 relation types for binding interaction.


\begin{table}[h!]
\centering
\caption{Overall results (\%) on test set of ADE. Two souces here include Wikidata and UMLS (restricted version)}
\vspace{2mm}
\begin{tabular}{@{\hskip 0.2in}c@{\hskip 0.3in}c@{\hskip 0.3in}c@{\hskip 0.2in}}
\hline
\textbf{Model} & \textbf{Entity
}& \textbf{Relation}\\
\hline
Relation-Metric (\cite{Tran2019-an}) & 87.11
& 77.29\\
SpERT (\cite{Eberts2019-xg}) & 89.28 & 78.84\\
SPAN$_{\text{\footnotesize{Multi-Head}}}$ (\cite{Ji2020-cl})& 90.59 & 80.73\\[1mm]
\hline
SentContextOnly & 88.13 & 77.23\\
FlatAttention & 89.16 & 78.81\\
KnowBertAttention & 90.08 & 79.95\\
KECI& \textbf{\color{blue}90.67} & 81.74\\
\hline
\textbf{Our Model (Wikidata)} &  \textbf{\color{red}90.92} & \textbf{\color{red}83.22}\\
\textbf{Our Model (UMLS)} & 90.58 & \textbf{\color{blue}81.86}\\
\hline
\end{tabular}
\label{res_ADE}
\end{table}

\subsection{Baselines}
We compare our method against state-of-the-art methods for both BioRelEx and ADE datasets. For evaluation, we use F-1 scores on entity and relation extraction for both BioRelEx and ADE. Baseline models include:
\begin{enumerate}
    \item Three previous models on entity and relation extraction: \textbf{Relation-Metric} \cite{Tran2019-an}, \textbf{SpERT \cite{Eberts2019-xg}}, and \textbf{SPAN$_{\text{\footnotesize{Multi-Head}}}$} \cite{Ji2020-cl}.
    \item \textbf{SentContextOnly:} This baseline uses only local context for prediction and does not use any external knowledge. It comes directly from the initial span graph construction step of KECI model.
    \item \textbf{FlatAttention:} This baseline does not use collective inference approach. As a result, the entity representation does not encode global relational information.
    \item \textbf{KnowBertAttention:} This baseline is a state-of-the-art knowledge-enhanced language model. Unlike KECI, it only uses SciBERT to extract feature from candidate entity mentions. 
    \item \textbf{KECI Full Model:} Instead of only extracting features as in KnowBertAttention, KECI makes use of multi-relational graph structure of candidate entities. By comparing our model with KECI, we investigate the potential for the use of a more generalized and reusable KG compared to task-dependent KG across datasets. 
\end{enumerate}

\subsection{Entities and relations extraction results}\label{res2}
\begin{figure*}[h!]
  \centering
  \includegraphics[scale=0.4]{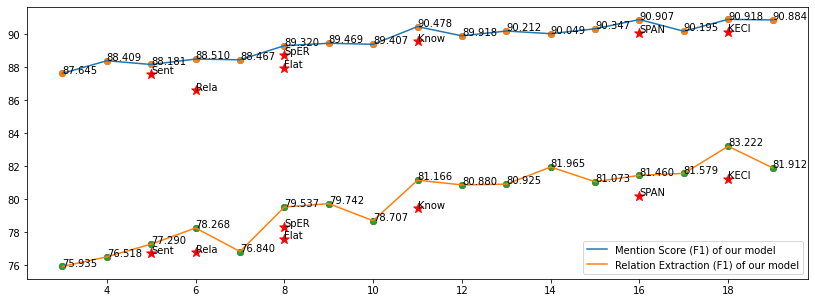}
  \caption{Testing results of ADE using our models with source Wikidata over 19 epochs.}
  \label{ade_wiki}
\end{figure*}

We train-test with 1-fold and provide the highest result within 50 epochs. \cref{res_ADE} shows results on test sets of ADE data for both sources Wikidata and UMLS. In addition, \cref{ade_wiki} shows the testing results of using Wikidata as source across 20 epochs and when it outperforms the baselines, and \cref{res_Bio} shows results on dev sets of BioRelEx with UMLS as source data in comparison to other baselines.\\

\begin{table}[h!]
\centering
\caption{Overall results (\%) on the development set of BioRelEx}
\vspace{2mm}
\begin{tabular}{@{\hskip 0.2in}c@{\hskip 0.3in}c@{\hskip 0.3in}c@{\hskip 0.2in}}
\hline
\textbf{Model}&\textbf{Entity
} & \textbf{Relation}\\
\hline
SciIE \cite{Luan2018-eb} & 77.90
& 49.60\\
\hline
SentContextOnly & 83.98 & 63.90\\
FlatAttention & 84.32 & 64.23\\
KnowBertAttention & 85.69 & \textbf{\color{blue}65.13}\\

KECI & \textbf{\color{red}87.35} & \textbf{\color{red}67.09}\\
\hline
\textbf{Our Model (UMLS)} & \textbf{\color{blue}87.31} & 63.21\\
\hline
\end{tabular}
\label{res_Bio}
\end{table}

\noindent Overall, we see that an improvement in runtime when running KECI framework versus our framework. This is expected as with our training process, we don't need to retrain a significant part of the knowledge graph when dealing with specific tasks. Moreover, our model outperforms all baselines for ADE data set (using both sources) and matches closely for BioRelEx. We remark that the size of two sources differ vastly (1200 distinct entities for Wikidata to 9726 distinct entities for UMLS), and these two datasets focus on two distinct subfields of the biomedical domain. As a result, our model's performance effectively demonstrates the power of our approach to reuse knowledge across tasks and its potential to be generalized to more tasks.

\section{Discussion}
\subsection{Relation weights extraction}
\cref{rel} presents the results on extracting relation weights and predicting relation for the first GK component of our framework. We note that only correct predictions will be used in the later step of training FFNN. In addition, since we are using MetaMap and UMLS to build the initial span graph, for Wikidata, we can only use entities where we can identify the CUID. As a result, the available entities from Wikidata greatly reduces compared to that of UMLS.\\

\begin{table}[h!]
\centering
\caption{Prediction results on triples (subject, relation, object) for two sources UMLS and Wikidata.}\label{rel}
\vspace{2mm}
\begin{tabular}{@{\hskip 0.2in}c@{\hskip 0.3in}c@{\hskip 0.3in}c}
\hline
\textbf{Statistics}&\textbf{Wikidata
} & \textbf{UMLS}\\
\hline
\small{Relations used} & \small{3 relations}
& \small{5 relations}\\
Number of \small{entities pairs} & \small{12000 pairs} & \small{40000 pairs}\\
\small{Correct predictions} & \small{9924 pairs} & \small{20k pairs}\\
\small{Distinct Entities with CUID} & \small{1200} & \small{7762}\\
\hline
\end{tabular}
\end{table}

\begin{figure}[h!]
  \centering
  \includegraphics[scale=0.7]{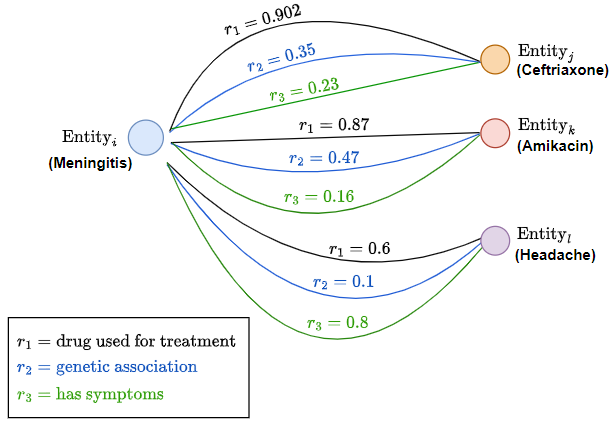}
  \caption{Example of relation weights for subject being disease meningitis.}
  \label{mening}
\end{figure}

\noindent Besides predicting relations (with associated weights), the framework also demonstrates that it can provide other related information. For example, \cref{mening} provides prediction weights for the subject $s_i$ \textbf{meningitis} (a disease), its two asssociated objects, which are drugs for the disease's treatment: $o_1$ = \textbf{ceftriaxone} and $o_2$ = \textbf{amikacin}, and finally the third object, which is the associated symptom $o_3$ = \textbf{headache}. Here, we observe that \textbf{ceftriaxone} has a higher weights than \textbf{amikacin} for being the appropriate drug used for the disease \textbf{meningitis}. This observation is validated by the fact that \textbf{ceftriaxone} is a more common antibiotics for \textbf{meningitis} than \textbf{amikacin}.

\subsection{Contribution of GK component}
We investigate how much re-usability the GK component contributes to the overall knowledge graph. To estimate such contribution, we count the number of distinct entities and the number of nodes/entities that frequently appear within the ADE dataset (nodes appearing in more than 10 sentences). \cref{count} shows how much the GK component built by Wikidata and UMLS contributes to the total number of entities in ADE. \\

\begin{table}[h!]
\centering
\caption{Graph statistics showing how many nodes GK allows for reusing across dataset. UMLS, which is bigger than Wikidata, contributes significantly to more frequent nodes.}
\vspace{2mm}
\begin{tabular}{@{\hskip 0.2in}c@{\hskip 0.3in}c@{\hskip 0.3in}c}
\hline
\textbf{Sources}&\textbf{Number of Distinct Nodes} & \textbf{Number of Frequent Nodes}\\
\hline
\small{ADE data} & \small{9947}
& \small{1206}\\
\small{Wikidata} & \small{391 (4\%)} & \small{91 (7.5\%)}\\
\small{UMLS} & \small{1401 (14\%)} & \small{247 (20\%)}\\
\hline
\end{tabular}
\label{count}
\end{table}

\noindent Our observations indicate that the GK component contributes more to the more-frequent entities within the dataset. We believe that contributing and interacting with frequent entities is one of the key points enabling our framework to produce matching results with other baselines without the need to train the whole GCN. Moreover, as we increase the size of the GK, the overall results stay quite stable and very close to those of KECI. This suggests that the relational embeddings obtained in \cref{part1} and \cref{part2} effectively capture relational information, performing comparably well compared to training a full GCN.\\

\noindent For BioRelEx dataset, we currently utilize a smaller and simplified subset of UMLS. This might leave out more important/frequent entities of this data set, and thus increasing the size of UMLS can help our approach closely match KECI model's performance on relation extraction for BioRelEx. Moreover, as shown in \cref{count}, approximately 20\% of nodes can already be reused, highlighting a significant amount of reusable information. Moreover, our framework achieves an accuracy on the ADE dataset that is only 0.09\% below that of using UMLS as a source for entity extraction, while outperming all other baselines in relation extraction. These observations further illustrate that our approach, which aim to build a universal and task-independent background knowledge, can be further extended to a larger scale to enhance its applicability and performance across diverse biomedical tasks.
\section{Conclusion and Future Works}
In this work, we introduce a knowledge-enhanced model supported by two components: GK (general-knowledge) and SK (specific-knowledge). We apply this model to different biomedical information extraction tasks and demonstrate competitive results on both ADE and BioRelEx datasets. The GK component of our framework serves as an effective general knowledge graph that proves reusable across multiple customized tasks. Experimental results on ADE and BioRelEx datasets also indicate potential scalability of our framework to broader applications. In the future, we plan to extend the framework in two directions. Firstly, we aim to expand on our GK component to improve both the prediction accuracy and also its re-usability on specific tasks. The second direction is to apply our framework to other research areas beyond biomedical fields such as literature, history, and architecture.\\

The implementation is available at \href{https://github.com/mpnguyen2/bio\_kg\_nlp}{https://github.com/mpnguyen2/bio\_kg\_nlp}.

\bibliographystyle{splncs04}
\bibliography{citation}

\end{document}